\pdfoutput=1

\documentclass[11pt]{article}

\usepackage{acl}

\usepackage{times}
\usepackage{latexsym}

\usepackage[T1]{fontenc}

\usepackage[utf8]{inputenc}

\usepackage{microtype}

\usepackage{inconsolata}

\usepackage{graphicx}
\usepackage{multirow}
\usepackage[skins]{tcolorbox}
\usepackage{subcaption}
\usepackage{xcolor}

\newtcolorbox{instructionbox}{
enhanced,
boxrule=0pt,frame hidden,
borderline west={4pt}{0pt}{Emerald!75!black},
colback=Emerald!9!white,
sharp corners
}

\newtcolorbox{problembox}{
enhanced,
boxrule=0pt,frame hidden,
borderline west={4pt}{0pt}{SkyBlue!55!black},
colback=SkyBlue!9!white,
sharp corners
}

\title{Culinary Class Wars: Evaluating LLMs using ASH \\ in Cuisine Transfer Task}

\author{
 \textbf{Hoonick Lee\textsuperscript{1}\thanks{Equal contributors}},
 \textbf{Mogan Gim\textsuperscript{2$*$}},
 \textbf{Donghyeon Park\textsuperscript{3}}
 \textbf{Donghee Choi\textsuperscript{4$\dagger$}},
 \textbf{Jaewoo Kang\textsuperscript{1}\thanks{Corresponding authors}}, 
 \\
 \textsuperscript{1}Korea University,
 \textsuperscript{2}Hankuk University of Foreign Studies,
 \\
 \textsuperscript{3}Sejong University,
 \textsuperscript{4}Imperial College London, \\
 \texttt{\{hoonick,kangj\}@korea.ac.kr},
 \texttt{gimmogan@hufs.ac.kr}, 
 \\
 \texttt{parkdh@sejong.ac.kr},
 \texttt{donghee.choi@imperial.ac.uk}
}

\begin{document}

\def\bench{$\mathcal{ASH}$~}
\def\aut{\colorbox{cyan}{\textbf{authenticity}}~}
\def\sen{\colorbox{yellow}{\textbf{sensitivity}}~}
\def\har{\colorbox{green}{\textbf{harmony}}~}

\definecolor{myblue}{RGB}{218,227,243}
\definecolor{mygreen}{RGB}{197,224,180}

\definecolor{Amethyst}{rgb}{0.6, 0.4, 0.8}
\definecolor{Emerald}{rgb}{0.31, 0.78, 0.47}
\definecolor{Grey}{rgb}{0.1, 0.1, 0.1}

\newcommand{\hn}[1]{{\color{Amethyst}\textbf{[Hoonick:} #1\textbf{]}}}
\newcommand{\cdh}[1]{{\color{red}\textbf{[Donghee:} #1\textbf{]}}}
\newcommand{\pd}[1]{{\color{orange}\textbf{[Donghyeon:} #1\textbf{]}}}
\newcommand{\mg}[1]{{\color{Emerald}\textbf{[Mogan:} #1\textbf{]}}}

\newenvironment{revision}[2][]
   {\vspace{0.2cm} \begin{tcolorbox}[breakable, enhanced, colback = yellow, 
   title = Revision \thepoint \  #2,#1,
   colbacktitle = red!85!black, colframe = red!75!black
   ]\normalfont}
   {\par\end{tcolorbox}}

\maketitle

\begin{abstract}
The advent of Large Language Models (LLMs) have shown promise in various creative domains, including culinary arts. However, many LLMs still struggle to deliver the desired level of culinary creativity, especially when tasked with adapting recipes to meet specific cultural requirements. This study focuses on \textbf{cuisine transfer}-applying elements of one cuisine to another-to assess LLMs' culinary creativity. We employ a diverse set of LLMs to generate and evaluate culturally adapted recipes, comparing their evaluations against LLM and human judgments. We introduce the $\mathcal{ASH}$ (\aut, \sen, \har) benchmark to evaluate LLMs' recipe generation abilities in the cuisine transfer task, assessing their cultural accuracy and creativity in the culinary domain. Our findings reveal crucial insights into both generative and evaluative capabilities of LLMs in the culinary domain, highlighting strengths and limitations in understanding and applying cultural nuances in recipe creation. The code and dataset used in this project is openly available in \url{https://github.com/dmis-lab/CulinaryASH/}.

\end{abstract}

\section{Introduction}
Culinarians around the globe are striving to innovate new recipe ideas. Such creative process involves exploring novel ingredient pairings~\citep{park2019kitchenette}, combinations~\citep{gim2022recipemind} or adjusting ingredient proportions~\citep{choi2023kitchenscale}. Meanwhile, cuisine is a cultural embodiment of culinary arts that can be associated with a specific region, religion, or history~\citep{kocevski2020defining,nguyen2022refined,nguyen2023extracting,palta-rudinger-2023-fork}. It defines the style of handling ingredients to create dishes typically recorded as recipes. As the culinary world is becoming increasingly culturally interconnected, \textbf{cuisine transfer}, which involves applying the elements of the \textit{source} cuisine to base recipe of \textit{target} cuisine, has become a well-established practice for creating new recipe possibilities~\citep{shin2024generative,markham2023foodfusion}. 

As cuisine transfer can be treated as a sub-task of textual recipe generation, we avert our focus towards the recent technical advancements in generative large language models (LLMs)~\citep{team2024gemma,touvron2023llama,dubey2024llama,jiang2023mistral,gpt4omini}. Conventional LLMs possess the ability to understand users' instruction-based prompts and generate desirable responses based on their massively parameterized knowledge~\citep{ouyang2022training,chung2024scaling,rafailov2024direct}. They are also capable of evaluating LLM-generated texts which helps relieve the burden of manual evaluation by humans~\cite{min-etal-2023-factscore}. These advancements have led to the advent of automatic evaluation frameworks which are now a well-established research topic.

\begin{table*}[htp!]
\centering
\scalebox{0.70}
{
\begin{tabular}{|lll|l|l|ll|}
\hline
\multicolumn{3}{|c|}{\textbf{Regional Cuisines~(30) }} & \multicolumn{1}{c|}{\textbf{Religious Cuisines~(6)}} & \multicolumn{1}{c|}{\textbf{Historical Cuisines~(4)}} & \multicolumn{2}{c|}{\textbf{Base Food Dishes~(20)}} \\ \hline
Algerian        & Peruvian       & British       & Buddhist                                         & Aztec                                            & Barbecued meat         & Pancake               \\
Egyptian        & Southern US    & French        & Hindu                                            & Byzantine                                        & Burger                 & Pasta                 \\

...        & & & ... & ... & ... & \\
\hline
\end{tabular}
}
\caption{Truncated List of cuisines and base food dishes used in this study with numbers indicating the total number of items in each category. The complete list can be found in the appendix, Table~\ref{tab:bench_cuisine_dishes_all}.}
\label{tab:bench_cuisine_dishes}
\end{table*}

Despite these circumstances, there have been seldom previous works that attempted to develop a LLM-driven evaluation framework especially in the culinary domain. This is mainly due to cooking being more related to \textit{creativity} rather than \textit{factuality} where the latter has been widely explored by previous works~\cite{jeong2024olaph, bishop-etal-2024-longdocfactscore, tang-etal-2024-minicheck, chiang2024mergingfacts, zheng2023judging,manakul-etal-2023-selfcheckgpt}. Evaluating generated recipes based on cooking creativity is a non-trivial task~\cite{choi-etal-2024-cookingsense, yagcioglu-etal-2018-recipeqa, liu2020multi, h2020recipegpt, chandu-etal-2019-storyboarding, le-etal-2023-improved,antognini-etal-2023-assistive} which presents a novel challenge when trying to employ automatic evaluation approaches in the culinary domain. It is also important to note that almost none of the automatic frameworks have been deployed in creativity-oriented generation tasks. 

To address this issue, we designed an automatic evaluation framework that comprises LLMs as recipe generators~\citep{h2020recipegpt} and evaluators. As creativity stems from knowledge~\citep{kenett2024role}, our aim is to comprehensively investigate both the generative and evaluation capabilities of currently released LLMs under a specific task called \textbf{cuisine transfer}. By analyzing the LLM-generated recipe texts and evaluation results, we derive implications of the current state of LLMs' culinary knowledge. A generator LLM with limited culinary creativity span would repeatedly use the cuisine-specific ingredients across different base recipes lacking variation. An evaluator LLM fixated towards a cuisine-specific ingredient, would assign high ratings to generated recipes simply for including it, regardless of its culinary essence. 

To the best of our knowledge, this is the first attempt to develop an evaluation framework tailored for the culinary domain. We constructed the \bench benchmark dataset and LLM generation-evaluation framework tailored to the cuisine transfer task. We propose a novel criteria \bench~to evaluate the generated recipes which are \aut, \sen and \har. We present our findings that were derived from in-depth investigation on both recipe generators and evaluators in this work.

\section{\bench~Benchmark}

\subsection{Creation of Cuisine Transfer Instructions}

To evaluate generative language models’ ability to adapt to diverse cuisine styles, we created 800 standardized cuisine transfer instructions based on 20 base dishes and 40 cuisines, expanding on prior work by incorporating religious and historical cuisines (e.g., Buddhist, Aztec) and achieving balanced continental representation. Table~\ref{tab:bench_cuisine_dishes} provides a summary of the cuisines and dishes used, with each prompt requesting a recipe that includes ingredients and step-by-step instructions. Details can be found in Supplementary Section~\ref{supp:creation_of_cuisine_transfer_instruction}

\subsection{Recipe Generation with Cuisine Transfer}
In the \bench benchmark, six open-source LLMs were each assigned 800 instructions for the cuisine transfer task, generating a total of 4,800 recipes. Details can be found in Supplementary Section~\ref{supp:recipe_generation_detail}.

\subsection{Evaluation of Generated Recipes}
\bench benchmark introduces the following evaluation criteria that measures a LLM's recipe geneeration ability in cuisine transfer tasks.
\begin{itemize}
    \item \aut: This criterion assesses how well the generated recipe maintains the essence of its original base dish. 
    \item \sen: This criterion assesses how well the generated recipe reflects the culinary elements transferred from the target cuisine.
    \item \har: This criterion not only assesses the overall quality of the generated recipe but also balance between \aut and \sen.
\end{itemize}

The details of evaluators are described in Supplementary section~\ref{supple:evaluator_detail}.

\section{Results}

\begin{figure*}[htbp]  %
    \centering
    \includegraphics[width=0.98\textwidth]{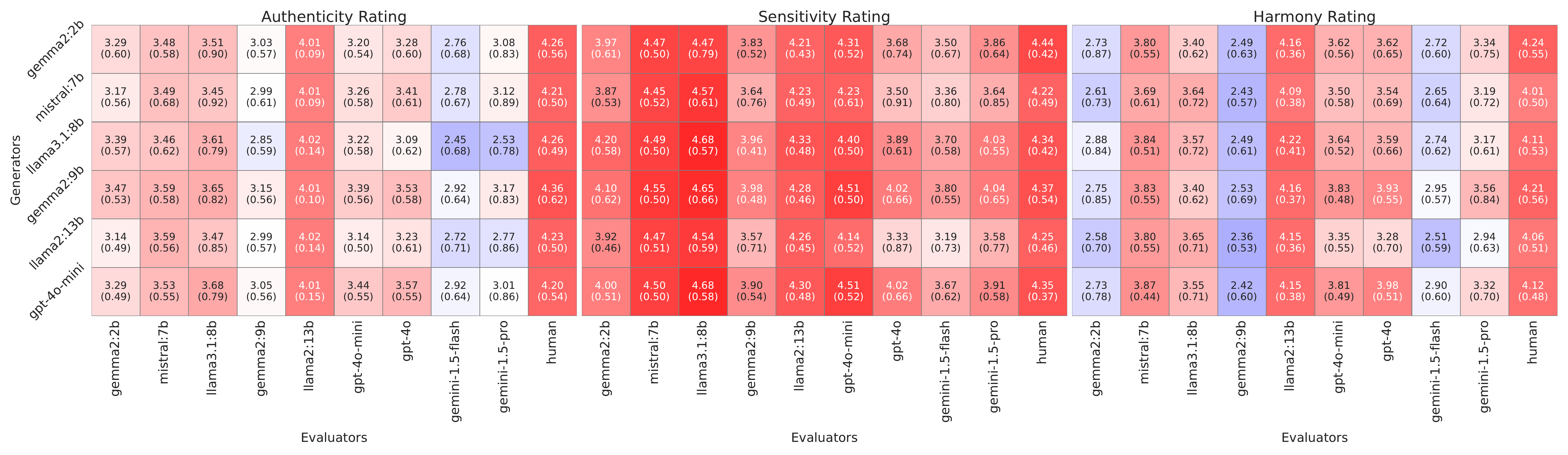}  %
    \caption{
    Mean and standard deviance of authenticity, sensitivity, harmony ratings calculated for each generator(y-axis)-evaluator(x-axis) pair across all cuisine transfers. The color scale ranges from 1 (blue) to 5 (red).}
    \label{fig:heatmap_scores}  %
\end{figure*}

\begin{table}[htp!]
\centering
\scalebox{0.68}
{
\centering
\begin{tabular}{|c|ccc|}
\hline
\textbf{Rank}                      & \textbf{\aut}                                                 & \textbf{\sen}                                                  & \textbf{\har}                                                      \\ \hline
\multirow{3}{*}{\textbf{Top 3}}    & \begin{tabular}[c]{@{}c@{}}Kosher \\ (3.959±0.573)\end{tabular}       & \begin{tabular}[c]{@{}c@{}}Kosher \\ (4.585±0.651)\end{tabular}       & \begin{tabular}[c]{@{}c@{}}Kosher \\ (3.786±0.861)\end{tabular}       \\ \cline{2-4} 
                                   & \begin{tabular}[c]{@{}c@{}}Islamic \\ (3.698±0.609)\end{tabular} & \begin{tabular}[c]{@{}c@{}}Islamic \\ (4.496±0.644)\end{tabular} & \begin{tabular}[c]{@{}c@{}}Islamic \\ (3.658±0.835)\end{tabular} \\ \cline{2-4} 
                                   & \begin{tabular}[c]{@{}c@{}}Canadian \\ (3.518±0.663)\end{tabular}     & \begin{tabular}[c]{@{}c@{}}Jain \\ (4.461±0.771)\end{tabular}    & \begin{tabular}[c]{@{}c@{}}Jain \\ (3.471±0.846)\end{tabular}    \\ \hline
$\cdots$                                & \multicolumn{3}{c|}{$\cdots$}                                                                                                                                                                                              \\ \hline
\multirow{3}{*}{\textbf{Bottom 3}} & \begin{tabular}[c]{@{}c@{}}Aztec \\ (3.145±0.722)\end{tabular}        & \begin{tabular}[c]{@{}c@{}}Brazilian \\ (3.865±0.79)\end{tabular}     & \begin{tabular}[c]{@{}c@{}}Zoroastrian \\ (3.21±0.812)\end{tabular}   \\ \cline{2-4} 
                                   & \begin{tabular}[c]{@{}c@{}}Chinese \\ (3.144±0.758)\end{tabular}      & \begin{tabular}[c]{@{}c@{}}Russian \\ (3.857±0.767)\end{tabular}      & \begin{tabular}[c]{@{}c@{}}Russian \\ (3.204±0.832)\end{tabular}      \\ \cline{2-4} 
                                   & \begin{tabular}[c]{@{}c@{}}Ethiopian \\ (3.101±0.744)\end{tabular}    & \begin{tabular}[c]{@{}c@{}}Zoroastrian \\ (3.739±0.883)\end{tabular}  & \begin{tabular}[c]{@{}c@{}}Ethiopian \\ (3.187±0.81)\end{tabular}     \\ \hline
\end{tabular}
}
\caption{Top and bottom 3 cuisines ranked by authenticity, sensitivity and harmony ratings expresed as mean and standard variance calculated across all generator-evaluator pairs} 
\label{tab:result_cuisine_ratings}
\end{table}

\begin{figure}[htbp]
    \centering
    \resizebox{0.95\linewidth}{!}{ %
        \begin{minipage}{\linewidth}
            \begin{subfigure}[]{\linewidth}
                \centering
                \includegraphics[width=\linewidth]{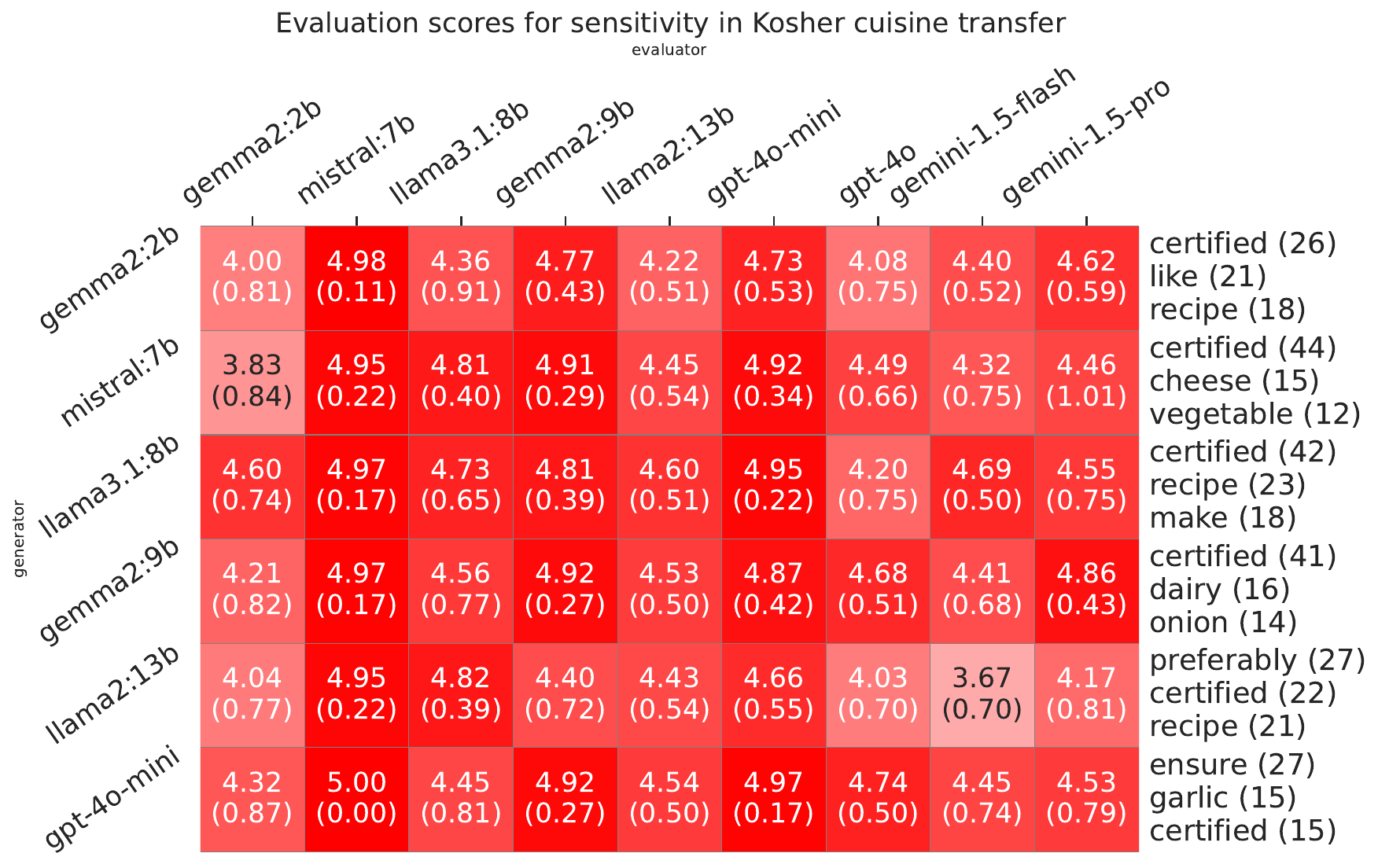}
                \caption{Sensitivity ratings and top frequently used words for Kosher cuisine transfer.}
                \label{fig:heatmap_wordcount_kosher}
            \end{subfigure}
            
            \begin{subfigure}[]{\linewidth}
                \centering
                \includegraphics[width=\linewidth]{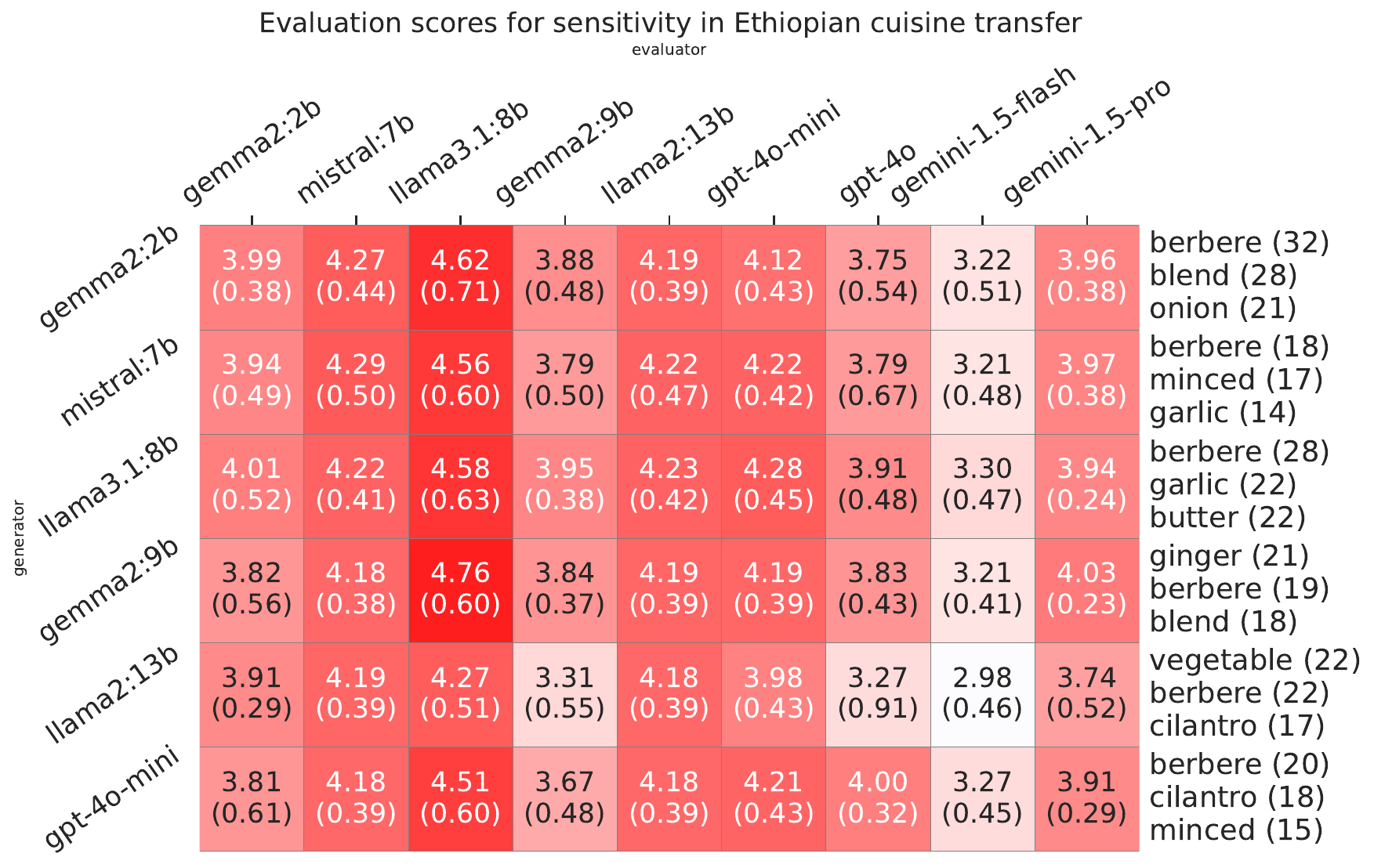}
                \caption{Sensitivity ratings and top frequently used words for Ethiopian cuisine transfer.}
                \label{fig:heatmap_wordcount_ethiopian}
            \end{subfigure}

            \begin{subfigure}[]{\linewidth}
                \centering
                \includegraphics[width=\linewidth]{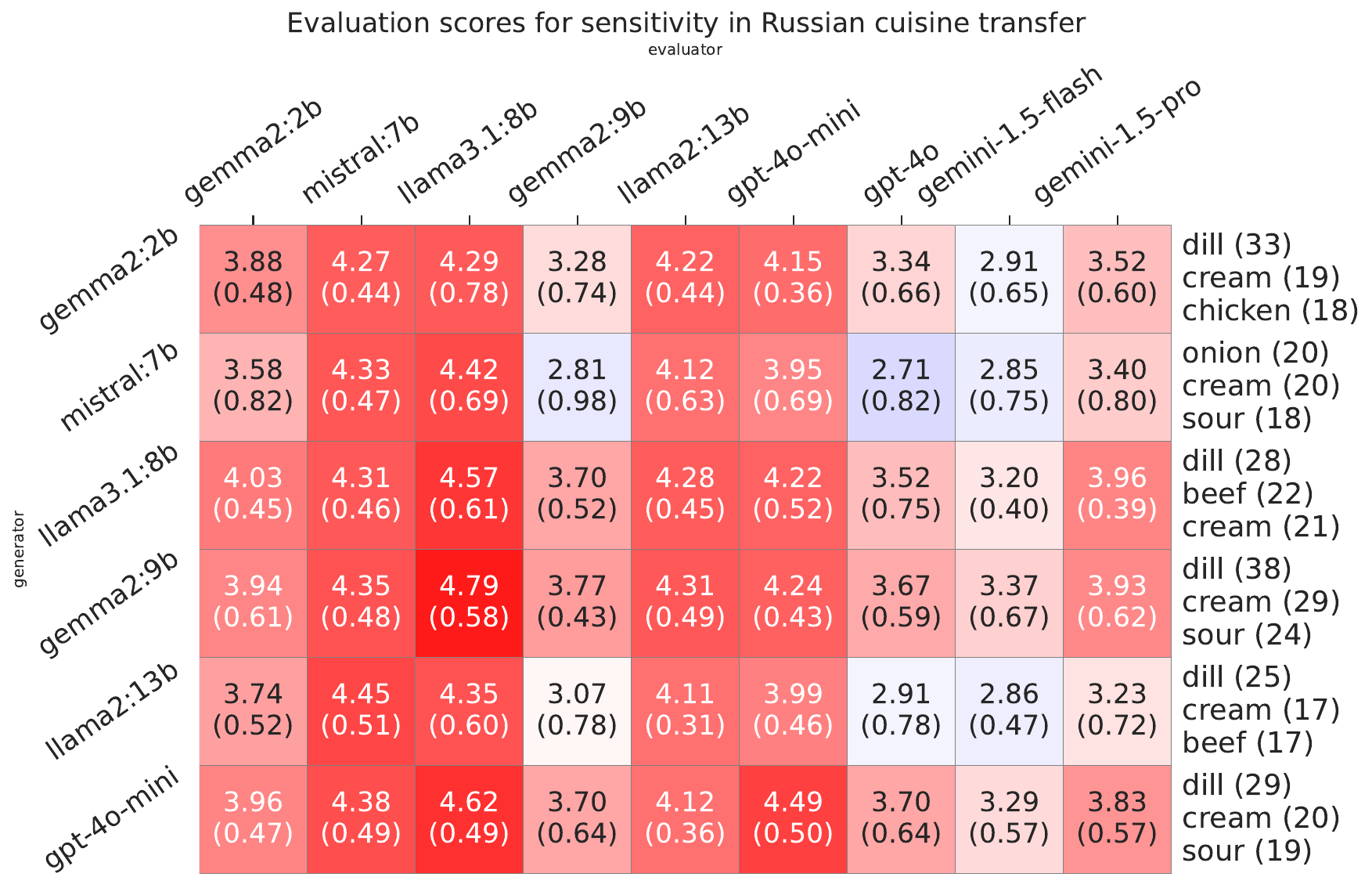}
                \caption{Sensitivity ratings and top frequently used words for Russian cuisine transfer.}
                \label{fig:heatmap_wordcount_russian}
            \end{subfigure}
        \end{minipage}
    }
    \caption{Sensitivity ratings and top 3 frequently used words for each cuisine transfer.}
    \label{fig:heatmap_wordcount}
\end{figure}

\subsection{Overall analysis of ratings assigned to LLM-generated recipes}
We generated heatmaps using the ratings of all evaluation criteria made by the LLM evaluators (Figure~\ref{fig:heatmap_scores}). 
While the distribution of ratings assigned to the six recipe generators were relatively even, we observed several differences across the recipe evaluators. In \aut, both \texttt{gemini} variants assigned relatively low ratings while \texttt{llama2:13b} was inclined towards all generators with the lowest deviance of its ratings. The ratings were harsher in \har where \texttt{gemma2:2b}, \texttt{gemma2:9b} and \texttt{gemini-1.5-flash} were generally more negative compared to others. In contrast, other evaluators displayed more leniency.

The overall distribution of ratings for \sen is way higher than the two evaluation criteria with 4.68 being the highest average rating assigned by \texttt{llama3.1:8b} to its identical model and 3.19 being the lowest assigned by \texttt{gemini-1.5-flash} to \texttt{llama2:13b}. We hypothesize that high \sen with relatively low \aut and \har means both the recipe generator and evaluator possess \textit{limited culinary knowledge}. The generator likely focuses heavily on incorporating specific culinary elements without considering the broader cooking context. Similarly, the evaluator may be overly focused on the presence of these elements in the text, rather than evaluating the overall coherence of the recipe.

\subsection{Cuisine-specific analysis of ratings assigned to LLM-generated recipes}

We investigated which cuisines have the highest and lowest ratings across all generator-evaluator LLM pairs for each evaluation criterion in the \bench benchmark experiment. Table~\ref{tab:result_cuisine_ratings} shows the top and bottom ranked cuisines for each criterion. Interestingly, recipes generated through Ethiopian cuisine transfer received the lowest ratings only in \aut and \har, while the \sen ratings are relatively ranked higher. 

For deeper investigation, we generated heatmaps (Figure~\ref{fig:heatmap_wordcount} using the \sen ratings provided by the recipe evaluators. 
We additionally examined the top frequently used words (excluding stop-words and cuisine names) for each generator model as well.

\begin{figure}[tbp]  %
    \centering
    \includegraphics[width=0.4\textwidth]{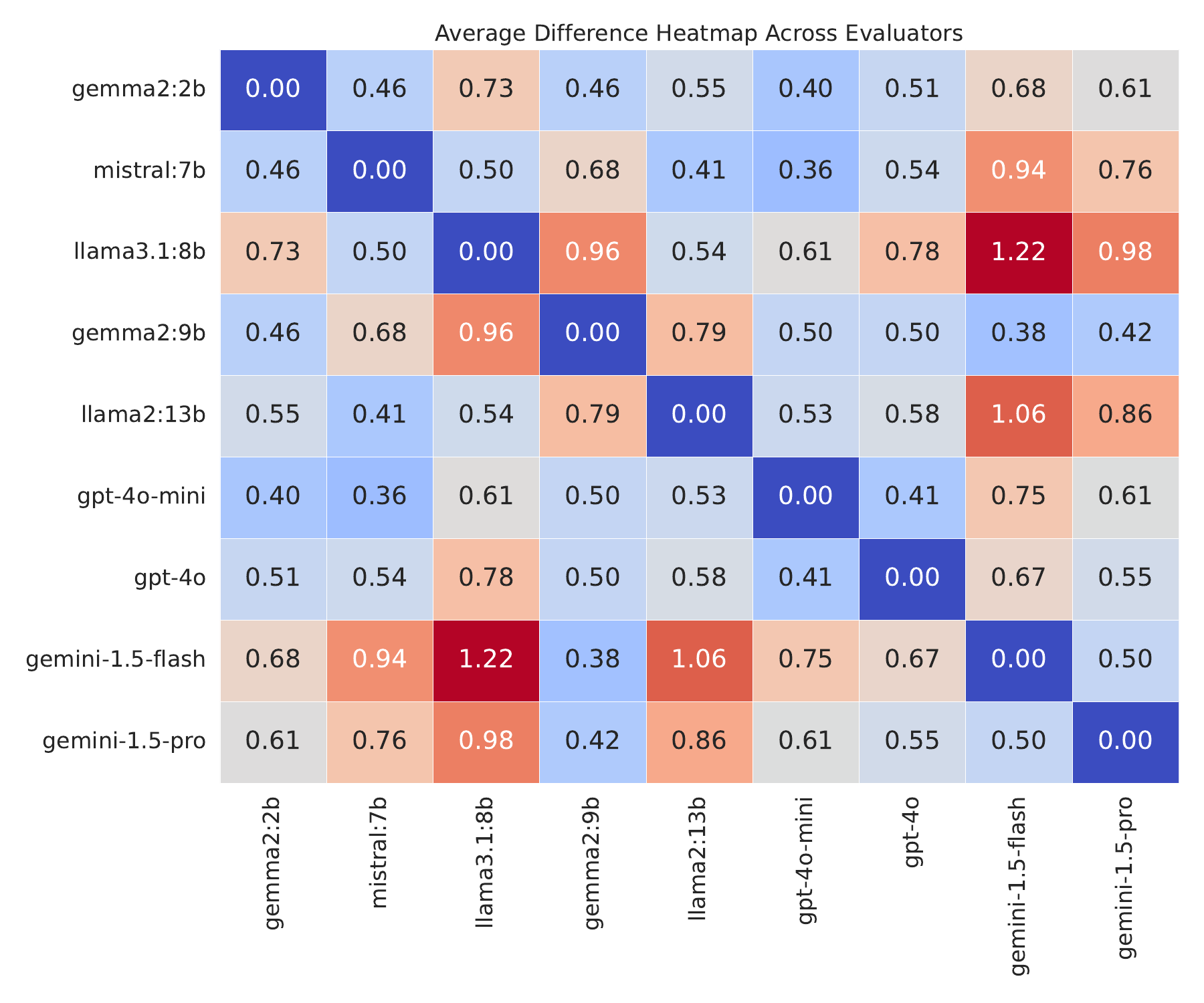}  %
    \caption{Average absolute difference values across evaluators.}
    \label{fig:average_diffrence}  %
\end{figure}

In Figure~\ref{fig:heatmap_wordcount_kosher}, all recipe generators received highly positive \sen ratings from almost all evaluators for the Kosher cuisine transfer task. As shown in Table~\ref{tab:result_cuisine_ratings}, Kosher cuisine remarkably achieved the highest ratings in all three criteria. In fact, one of the mostly used words in the generated recipes is \textit{certified} which may implicate Kosher-certified food ingredients. After examining the generated recipes and evaluation rationales, we discovered that the generators' focus tended to be polarized towards \textit{certified}. Conclusively, the strict emphasis on this aspect appeared to drive more favorable ratings from the evaluators.

According to Figure~\ref{fig:heatmap_wordcount_ethiopian}, one of the most frequently used words is \textit{berbere}, a spice mixture popularly used in Ethopian dishes. While this ingredient played a pivotal role in attracting positive \sen ratings, we speculate that its combination with the base dishes did not seem to help the generator earn good \aut and \har ratings from the evaluators which may be due to the unexplored culinary knowledge space related to Ethiopian cuisine in conventional LLMs. 

In Figure~\ref{fig:heatmap_wordcount_russian}, recipe generators \texttt{mistral:7b} and \texttt{llama2:13b} obtained relatively negative ratings especially from \texttt{gemma2:9b}, \texttt{gpt-4o} and \texttt{gemini-1.5-pro}.

\subsection{Inter-agreement analysis of recipe evaluators}
\begin{table}[]
\scalebox{0.73}
{
\begin{tabular}{|l|ccc|}
\hline
\textbf{LLM}     & \aut & \sen & \har \\ \hline
gemma2:2b        & 0.992                 & 0.591                & 1.464            \\ \hline
mistral:7b       & 0.853                 & 0.595                & 0.577            \\ \hline
llama3.1:8b      & 0.871                 & 0.685                & 0.915            \\ \hline
gemma2:9b        & 1.290                 & 0.672                & 1.760            \\ \hline
llama2:13b       & 0.503                 & 0.550                & 0.460            \\ \hline
gpt-4o-mini      & 1.085                 & 0.556                & 0.728            \\ \hline
gpt-4o           & 1.043                 & 0.819                & 0.762            \\ \hline
gemini-1.5-flash & 1.552                 & 0.929                & 1.488            \\ \hline
gemini-1.5-pro   & 1.370                 & 0.739                & 1.137            \\ \hline
\end{tabular}
}
\caption{Average rating differences between LLM and human annotators for each LLM recipe evaluator and evaluation rating. Lower values mean higher inter-agreement.}
\label{tab:result_human_llm_agreement}
\end{table}

We performed two inter-agreement analysis of recipe evaluators which are LLM-LLM and LLM-human annotators. Figure~\ref{fig:heatmap_agree} shows the LLM-LLM inter-agreement scores averaged across the evaluation criteria. While the \texttt{gemini} and \texttt{llama} variants seem to have discrepancy to each other, LLMs from the same family tend to resemble each other in terms of their agreement scores. As shown in~\ref{tab:result_human_llm_agreement}, we observed that \texttt{llama2:13b} tended to show the least rating differences in all evaluation criteria (0.503, 0.550, 0.460). Considering that the human annotators are not culinary experts, we speculate that evaluators with limited culinary knowledge may possess narrower recipe evaluation standards, often assigning more positive ratings, a trend also observed in the human evaluations.

\section{Conclusion}
We performed an thorough investigation on LLMs' generative behavior under a specific culinary domain task called cuisine transfer. We developed a novel benchmark $\mathcal{ASH}$ that addresses the important aspects of cuisine transfer when evaluating the generated recipes using \aut, \sen and \har. We analyzed the evaluation results yielded by our designed automatic evaluation framework and derived interesting findings that hints the current state of culinary knowledge in LLMs for future research in this creativity domain.

\section{Ethical Statement}

Our study examines the ability of large language models (LLMs) to generate culturally sensitive recipe adaptations through cuisine transfer, which involves incorporating elements from one cuisine into another. We are aware that LLMs may sometimes produce outputs that could inadvertently misrepresent or oversimplify cultural practices. To mitigate this, we have implemented evaluation criteria that prioritize authenticity, sensitivity, and harmony, aiming to ensure that generated recipes are respectful and reflective of the intended cultural elements.
However, we acknowledge that despite our efforts, the model may still generate content that could be viewed as controversial or culturally insensitive, depending on individual perspectives.

Additionally, we recognize the importance of human input in evaluating cultural nuances, so we involved a diverse group of human annotators from various cultural backgrounds. While the range of human raters was limited, we strived for as much diversity as possible to provide balanced perspectives. All data used in this study, including recipes and prompts, were sourced from publicly available datasets. This research does not involve any personal or sensitive data, and all annotations were conducted by adults who consented to participate in the evaluation.

\section{Limitations}

Our study has several limitations, primarily related to the constraints of using LLMs for nuanced cultural tasks. First, while we calculated agreement scores to assess the alignment between LLM and human evaluations, traditional Kappa scores are not ideal for continuous score variations. Using regression-based approaches might yield more accurate measures of evaluator agreement, and further research could explore these alternatives.

The results may also be vulnerable to variations in prompt setup and evaluation format. Prompt design significantly affects model output, and subtle differences in phrasing can lead to diverse interpretations by the LLMs. Similarly, the rating scale we used may oversimplify complex culinary judgments, potentially limiting the depth of assessment for each criterion. Future work could involve refining prompts and evaluation metrics for more nuanced feedback.

The study is additionally limited by the dependency on specific models, particularly proprietary models like GPT-4, which may introduce biases based on their training data and inaccessible architectural details. These model-specific factors mean that results may vary with other LLMs or future iterations of the same models.

Finally, our use of human annotators, while culturally diverse, is not exhaustive. The sample size and diversity of annotators could be expanded in future studies to encompass a wider range of cultural perspectives. This would further enrich our understanding of cultural sensitivities and improve the robustness of our evaluation framework.

\section{Potential Risks}

While our dataset is designed to evaluate culturally sensitive recipe generation, if it is used to further train or fine-tune language models, it may lead to unintended behaviors. This is due to the inherent variability in generated text, as well as potential limitations in the suitability of human or machine evaluations for certain cultural contexts.

Despite our efforts to ensure dataset quality, the limited number and cultural diversity of evaluators may mean our dataset does not comprehensively represent all perspectives within the cultural and culinary domains. This could restrict its effectiveness in capturing the full breadth of cultural nuance necessary for such applications.

\bibliography{9_references.bib,anthology}

\section{Appendix}
\appendix

\section{Experiment setup}
For each generator-evaluator pair in each evaluation criterion (\textbf{authenticity}, \textbf{sensitivity}, \textbf{harmony}), we calculated the mean and standard deviation of the 4,000 ratings across all generated recipes which underwent cuisine transfer. 

\begin{table}[]
\centering
\scalebox{0.6}
{
\begin{tabular}{|c|c|c|}
\hline
\textbf{\aut}                                                 & \textbf{\sen}                                                  & \textbf{\har}                                                      \\ \hline
\begin{tabular}[c]{@{}c@{}}Kosher \\ (3.959±0.573)\end{tabular}       & \begin{tabular}[c]{@{}c@{}}Kosher \\ (4.585±0.651)\end{tabular}       & \begin{tabular}[c]{@{}c@{}}Kosher \\ (3.786±0.861)\end{tabular}       \\ \hline
\begin{tabular}[c]{@{}c@{}}Islamic \\ (3.698±0.609)\end{tabular} & \begin{tabular}[c]{@{}c@{}}Islamic \\ (4.496±0.644)\end{tabular} & \begin{tabular}[c]{@{}c@{}}Islamic \\ (3.658±0.835)\end{tabular} \\ \hline
\begin{tabular}[c]{@{}c@{}}Canadian \\ (3.518±0.663)\end{tabular}     & \begin{tabular}[c]{@{}c@{}}Jain \\ (4.461±0.771)\end{tabular}    & \begin{tabular}[c]{@{}c@{}}Jain \\ (3.471±0.846)\end{tabular}    \\ \hline
\begin{tabular}[c]{@{}c@{}}Australian \\ (3.486±0.628)\end{tabular}   & \begin{tabular}[c]{@{}c@{}}Korean \\ (4.299±0.591)\end{tabular}       & \begin{tabular}[c]{@{}c@{}}Mexican \\ (3.469±0.832)\end{tabular}      \\ \hline
\begin{tabular}[c]{@{}c@{}}Southern US \\ (3.476±0.732)\end{tabular}  & \begin{tabular}[c]{@{}c@{}}Southern US \\ (4.245±0.637)\end{tabular}  & \begin{tabular}[c]{@{}c@{}}Korean \\ (3.462±0.84)\end{tabular}        \\ \hline
\begin{tabular}[c]{@{}c@{}}French \\ (3.466±0.735)\end{tabular}       & \begin{tabular}[c]{@{}c@{}}Greek \\ (4.241±0.584)\end{tabular}        & \begin{tabular}[c]{@{}c@{}}Southern US \\ (3.461±0.841)\end{tabular}  \\ \hline
\begin{tabular}[c]{@{}c@{}}Spanish \\ (3.455±0.657)\end{tabular}      & \begin{tabular}[c]{@{}c@{}}Mexican \\ (4.218±0.537)\end{tabular}      & \begin{tabular}[c]{@{}c@{}}Greek \\ (3.458±0.867)\end{tabular}        \\ \hline
\begin{tabular}[c]{@{}c@{}}British \\ (3.444±0.703)\end{tabular}      & \begin{tabular}[c]{@{}c@{}}Japanese \\ (4.207±0.568)\end{tabular}     & \begin{tabular}[c]{@{}c@{}}Italian \\ (3.452±0.896)\end{tabular}      \\ \hline
\begin{tabular}[c]{@{}c@{}}Greek \\ (3.442±0.694)\end{tabular}        & \begin{tabular}[c]{@{}c@{}}Italian \\ (4.203±0.655)\end{tabular}      & \begin{tabular}[c]{@{}c@{}}Indian \\ (3.416±0.806)\end{tabular}       \\ \hline
\begin{tabular}[c]{@{}c@{}}Italian \\ (3.422±0.79)\end{tabular}       & \begin{tabular}[c]{@{}c@{}}Hindu diet \\ (4.198±0.615)\end{tabular}   & \begin{tabular}[c]{@{}c@{}}Buddhist \\ (3.413±0.859)\end{tabular}     \\ \hline
\begin{tabular}[c]{@{}c@{}}Irish \\ (3.415±0.649)\end{tabular}        & \begin{tabular}[c]{@{}c@{}}Thai \\ (4.191±0.524)\end{tabular}         & \begin{tabular}[c]{@{}c@{}}Japanese \\ (3.411±0.822)\end{tabular}     \\ \hline
\begin{tabular}[c]{@{}c@{}}Jamaican \\ (3.396±0.657)\end{tabular}     & \begin{tabular}[c]{@{}c@{}}Indian \\ (4.188±0.521)\end{tabular}       & \begin{tabular}[c]{@{}c@{}}Spanish \\ (3.411±0.846)\end{tabular}      \\ \hline
\begin{tabular}[c]{@{}c@{}}Filipino \\ (3.387±0.682)\end{tabular}     & \begin{tabular}[c]{@{}c@{}}Jamaican \\ (4.172±0.576)\end{tabular}     & \begin{tabular}[c]{@{}c@{}}French \\ (3.408±0.874)\end{tabular}       \\ \hline
\begin{tabular}[c]{@{}c@{}}Korean \\ (3.38±0.701)\end{tabular}        & \begin{tabular}[c]{@{}c@{}}Filipino \\ (4.171±0.649)\end{tabular}     & \begin{tabular}[c]{@{}c@{}}Canadian \\ (3.403±0.831)\end{tabular}     \\ \hline
\begin{tabular}[c]{@{}c@{}}Peruvian \\ (3.365±0.672)\end{tabular}     & \begin{tabular}[c]{@{}c@{}}Vietnamese \\ (4.131±0.646)\end{tabular}   & \begin{tabular}[c]{@{}c@{}}Thai \\ (3.4±0.819)\end{tabular}           \\ \hline
\begin{tabular}[c]{@{}c@{}}Hawaiian \\ (3.359±0.681)\end{tabular}     & \begin{tabular}[c]{@{}c@{}}Buddhist \\ (4.121±0.802)\end{tabular}     & \begin{tabular}[c]{@{}c@{}}Jamaican \\ (3.387±0.817)\end{tabular}     \\ \hline
\begin{tabular}[c]{@{}c@{}}Buddhist \\ (3.341±0.723)\end{tabular}     & \begin{tabular}[c]{@{}c@{}}Moroccan \\ (4.111±0.525)\end{tabular}     & \begin{tabular}[c]{@{}c@{}}Ottoman \\ (3.382±0.799)\end{tabular}      \\ \hline
\begin{tabular}[c]{@{}c@{}}Japanese \\ (3.335±0.708)\end{tabular}     & \begin{tabular}[c]{@{}c@{}}Spanish \\ (4.104±0.697)\end{tabular}      & \begin{tabular}[c]{@{}c@{}}Filipino \\ (3.376±0.82)\end{tabular}      \\ \hline
\begin{tabular}[c]{@{}c@{}}Algerian \\ (3.332±0.671)\end{tabular}     & \begin{tabular}[c]{@{}c@{}}Peruvian \\ (4.09±0.633)\end{tabular}      & \begin{tabular}[c]{@{}c@{}}Moroccan \\ (3.372±0.818)\end{tabular}     \\ \hline
\begin{tabular}[c]{@{}c@{}}Mexican \\ (3.33±0.685)\end{tabular}       & \begin{tabular}[c]{@{}c@{}}Chinese \\ (4.086±0.543)\end{tabular}      & \begin{tabular}[c]{@{}c@{}}Hindu diet \\ (3.365±0.815)\end{tabular}   \\ \hline
\begin{tabular}[c]{@{}c@{}}Costa Rican \\ (3.325±0.691)\end{tabular}  & \begin{tabular}[c]{@{}c@{}}French \\ (4.077±0.767)\end{tabular}       & \begin{tabular}[c]{@{}c@{}}Peruvian \\ (3.361±0.829)\end{tabular}     \\ \hline
\begin{tabular}[c]{@{}c@{}}Brazilian \\ (3.325±0.674)\end{tabular}    & \begin{tabular}[c]{@{}c@{}}Hawaiian \\ (4.077±0.682)\end{tabular}     & \begin{tabular}[c]{@{}c@{}}Vietnamese \\ (3.341±0.838)\end{tabular}   \\ \hline
\begin{tabular}[c]{@{}c@{}}Scottish \\ (3.312±0.707)\end{tabular}     & \begin{tabular}[c]{@{}c@{}}Canadian \\ (4.07±0.789)\end{tabular}      & \begin{tabular}[c]{@{}c@{}}Hawaiian \\ (3.337±0.82)\end{tabular}      \\ \hline
\begin{tabular}[c]{@{}c@{}}Swedish \\ (3.306±0.712)\end{tabular}      & \begin{tabular}[c]{@{}c@{}}Ottoman \\ (4.069±0.593)\end{tabular}      & \begin{tabular}[c]{@{}c@{}}Irish \\ (3.325±0.786)\end{tabular}        \\ \hline
\begin{tabular}[c]{@{}c@{}}Ottoman \\ (3.296±0.663)\end{tabular}      & \begin{tabular}[c]{@{}c@{}}Scottish \\ (4.02±0.693)\end{tabular}      & \begin{tabular}[c]{@{}c@{}}Algerian \\ (3.323±0.81)\end{tabular}      \\ \hline
\begin{tabular}[c]{@{}c@{}}Thai \\ (3.292±0.706)\end{tabular}         & \begin{tabular}[c]{@{}c@{}}Algerian \\ (4.002±0.621)\end{tabular}     & \begin{tabular}[c]{@{}c@{}}Australian \\ (3.323±0.81)\end{tabular}    \\ \hline
\begin{tabular}[c]{@{}c@{}}Vietnamese \\ (3.291±0.762)\end{tabular}   & \begin{tabular}[c]{@{}c@{}}Ethiopian \\ (3.994±0.57)\end{tabular}     & \begin{tabular}[c]{@{}c@{}}British \\ (3.317±0.849)\end{tabular}      \\ \hline
\begin{tabular}[c]{@{}c@{}}Zoroastrian \\ (3.287±0.747)\end{tabular}  & \begin{tabular}[c]{@{}c@{}}Irish \\ (3.982±0.735)\end{tabular}        & \begin{tabular}[c]{@{}c@{}}Chinese \\ (3.308±0.81)\end{tabular}       \\ \hline
\begin{tabular}[c]{@{}c@{}}Moroccan \\ (3.277±0.663)\end{tabular}     & \begin{tabular}[c]{@{}c@{}}British \\ (3.97±0.831)\end{tabular}       & \begin{tabular}[c]{@{}c@{}}Scottish \\ (3.269±0.775)\end{tabular}     \\ \hline
\begin{tabular}[c]{@{}c@{}}Indian \\ (3.269±0.68)\end{tabular}        & \begin{tabular}[c]{@{}c@{}}Swedish \\ (3.966±0.772)\end{tabular}      & \begin{tabular}[c]{@{}c@{}}Swedish \\ (3.268±0.845)\end{tabular}      \\ \hline
\begin{tabular}[c]{@{}c@{}}Egyptian \\ (3.267±0.683)\end{tabular}     & \begin{tabular}[c]{@{}c@{}}Polynesian \\ (3.947±0.615)\end{tabular}   & \begin{tabular}[c]{@{}c@{}}Costa Rican \\ (3.26±0.824)\end{tabular}   \\ \hline
\begin{tabular}[c]{@{}c@{}}Russian \\ (3.263±0.708)\end{tabular}      & \begin{tabular}[c]{@{}c@{}}Costa Rican \\ (3.942±0.723)\end{tabular}  & \begin{tabular}[c]{@{}c@{}}Polynesian \\ (3.259±0.801)\end{tabular}   \\ \hline
\begin{tabular}[c]{@{}c@{}}Jain \\ (3.26±0.765)\end{tabular}     & \begin{tabular}[c]{@{}c@{}}Byzantine \\ (3.918±0.657)\end{tabular}    & \begin{tabular}[c]{@{}c@{}}Byzantine \\ (3.257±0.795)\end{tabular}    \\ \hline
\begin{tabular}[c]{@{}c@{}}Medieval \\ (3.249±0.762)\end{tabular}     & \begin{tabular}[c]{@{}c@{}}Australian \\ (3.913±0.815)\end{tabular}   & \begin{tabular}[c]{@{}c@{}}Aztec \\ (3.253±0.787)\end{tabular}        \\ \hline
\begin{tabular}[c]{@{}c@{}}Hindu diet \\ (3.204±0.736)\end{tabular}   & \begin{tabular}[c]{@{}c@{}}Aztec \\ (3.913±0.673)\end{tabular}        & \begin{tabular}[c]{@{}c@{}}Egyptian \\ (3.239±0.809)\end{tabular}     \\ \hline
\begin{tabular}[c]{@{}c@{}}Byzantine \\ (3.194±0.726)\end{tabular}    & \begin{tabular}[c]{@{}c@{}}Medieval \\ (3.902±0.768)\end{tabular}     & \begin{tabular}[c]{@{}c@{}}Brazilian \\ (3.222±0.819)\end{tabular}    \\ \hline
\begin{tabular}[c]{@{}c@{}}Polynesian \\ (3.177±0.694)\end{tabular}   & \begin{tabular}[c]{@{}c@{}}Egyptian \\ (3.9±0.71)\end{tabular}        & \begin{tabular}[c]{@{}c@{}}Medieval \\ (3.222±0.789)\end{tabular}     \\ \hline
\begin{tabular}[c]{@{}c@{}}Aztec \\ (3.145±0.722)\end{tabular}        & \begin{tabular}[c]{@{}c@{}}Brazilian \\ (3.865±0.79)\end{tabular}     & \begin{tabular}[c]{@{}c@{}}Zoroastrian \\ (3.21±0.812)\end{tabular}   \\ \hline
\begin{tabular}[c]{@{}c@{}}Chinese \\ (3.144±0.758)\end{tabular}      & \begin{tabular}[c]{@{}c@{}}Russian \\ (3.857±0.767)\end{tabular}      & \begin{tabular}[c]{@{}c@{}}Russian \\ (3.204±0.832)\end{tabular}      \\ \hline
\begin{tabular}[c]{@{}c@{}}Ethiopian \\ (3.101±0.744)\end{tabular}    & \begin{tabular}[c]{@{}c@{}}Zoroastrian \\ (3.739±0.883)\end{tabular}  & \begin{tabular}[c]{@{}c@{}}Ethiopian \\ (3.187±0.81)\end{tabular}     \\ \hline
\end{tabular}
}
\caption{40 cuisines ranked by authenticity, sensitivity and harmony ratings expresed as mean and standard variance calculated across all generator-evaluator pairs}
\label{tab:my-table}
\end{table}

\begin{table*}[htp!]
\centering
\scalebox{0.75}
{
\begin{tabular}{|lll|l|l|ll|}
\hline
\multicolumn{3}{|c|}{\textbf{Regional Cuisines~(30) }} & \multicolumn{1}{c|}{\textbf{Religious Cuisines~(6)}} & \multicolumn{1}{c|}{\textbf{Historical Cuisines~(4)}} & \multicolumn{2}{c|}{\textbf{Base Food Dishes~(20)}} \\ \hline
Algerian        & Peruvian       & British       & Buddhist                                         & Aztec                                            & Barbecued meat         & Pancake               \\
Egyptian        & Southern US    & French        & Hindu                                            & Byzantine                                        & Burger                 & Pasta                 \\
Ethiopian        & Chinese        & Greek         & Islamic                                          & Medieval                                         & Burritos               & Pizza                 \\
Moroccan       & Filipino       & Irish         & Jain                                             & Ottoman                                          & Crepes                 & Rolls                 \\
Brazilian       & Indian         & Italian       & Kosher                                           &                                                  & Curry                  & Salad                 \\
Canadian        & Japanese       & Scottish      & Zoroastrian                                      &                                                  & Fried Chicken          & Sandwich              \\
Costa Rican     & Korean         & Spanish       &                                                  &                                                  & Fried Noodles          & Savory Waffle         \\
Hawaiian        & Russian        & Swedish       &                                                  &                                                  & Fried Rice             & Savoury Pie           \\
Jamaican        & Thai           & Australian    &                                                  &                                                  & French Fries           & Soup Noodle           \\
Mexican         & Vietnamese     & Polynesian    &                                                  &                                                  & Lasagna                & Stew                  \\ \hline
\end{tabular}
}
\caption{List of cuisines and base food dishes used in this study.}
\label{tab:bench_cuisine_dishes_all}
\end{table*}

\section{Creation of Cuisine Transfer Instructions Details}
\label{supp:creation_of_cuisine_transfer_instruction}

We first created cuisine transfer instructions to assess a generative language model's ability to understand and adapt to other cuisine styles. Inspired from~\citeauthor{shin2024generative}, we selected 20 base food dishes used in their work. 

Additionally, we incorporated the 20 cuisines used in their study, primarily categorized by region. As our benchmark aims to provide a comprehensive measure of an LLM's culinary awareness from a broader cultural perspective, we expanded the set to include religious and historical cuisines such as Buddhist and Aztec cuisines. Also, as regional cuisine can be clustered into continental categories, we added regional cuisines to achieve a balance across different continents. 

Table~\ref{tab:bench_cuisine_dishes} presents a truncated list of cuisines and base food dishes used in the benchmark. With the total number of cuisines and base food dishes being 40 and 20, we created 800 cuisine transfer instructions. Each prompt follows a standardized format specifying the base dish to be applied with a specific cuisine transfer and asking the model to generate a textual recipe comprising ingredients and instruction steps.

\section{Recipe Generation Details}
\label{supp:recipe_generation_detail}
In our proposed benchmark, 6 open-source LLMs (\texttt{gemma2:2b}~\citep{team2024gemma}, \texttt{gemma2:9b}~\citep{team2024gemma}, \texttt{llama2:13b}~\citep{touvron2023llama}, \texttt{llama3.1:8b}~\citep{dubey2024llama}, \texttt{mistral:7b}~\citep{jiang2023mistral}, \texttt{gpt-4o-mini}~\citep{gpt4omini}) were selected to perform the cuisine transfer task, each given with 800 instructions. The selection of models were based on their model size, as well as their popularity displayed in the Ollama\footnote{https://github.com/ollama/ollama} models catalog page. We utilized Ollama, an open-source platform for running the LLMs on the cuisine transfer task in a local environment. As a result, we collected 800 textual recipes comprising a list of ingredients and cooking instructions generated by each LLM which makes a total number of 4,800 recipes.

All LLM recipe generators were prompted with the same instruction format as follows,
\begin{center}
\fbox{\begin{minipage}[t]{0.98\linewidth}
\texttt{Can you apply the elements of <cuisine> to this dish and make it into a recipe?} \\
\texttt{Dish: <base recipe dish>}. \\
\texttt{The response should be in the following form for ingredients and instructions each. For example:} \\
\texttt{ingredients:} \\
\texttt{<ingredient1>} \\
\texttt{<ingredient2>} \\
\texttt{...} \\
\texttt{instructions:} \\
\texttt{<instruction1>} \\
\texttt{<instruction2>} \\
\texttt{...} 
\end{minipage}}
\end{center}

\begin{figure*}[htbp]  %
    \centering
    \includegraphics[width=0.98\textwidth]{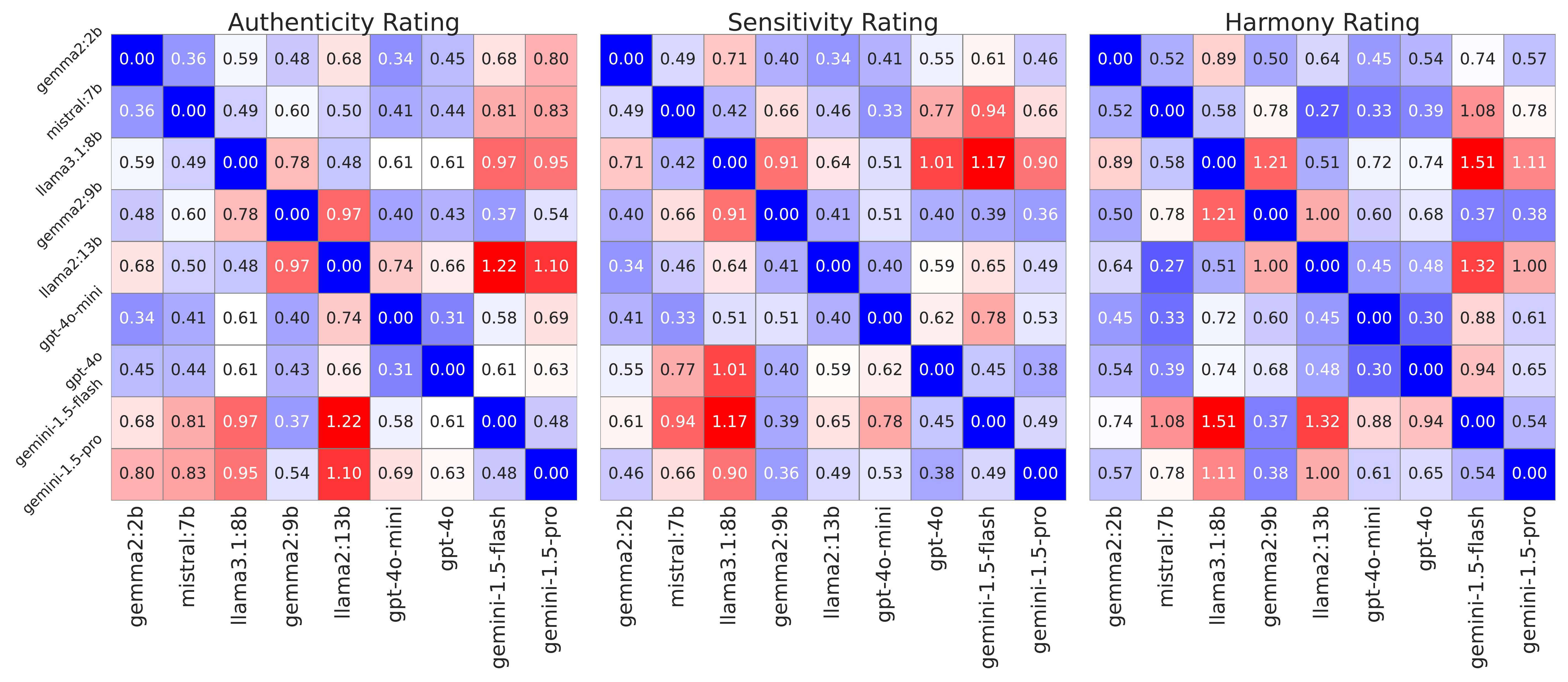}  %
    \caption{
    Average evaluation scores of authenticity, sensitivity, and harmony ratings were calculated for each generator(y-axis)-evaluator(x-axis) pair across all cuisine transfers. The color scale ranges from 0 (blue) to 1 (red).}
    \label{fig:heatmap_agree}  %
\end{figure*}

\section{Recipe Evaluation Details}

\begin{figure*}
\fbox{\begin{minipage}[t]{0.98\linewidth}
\texttt{Evaluate the following recipe:} \\
\texttt{Original Dish: <base food dish>} \\
\texttt{Variation: <cuisine>} \\
\texttt{Generated Recipe: <generated recipe>} \\
\texttt{Please rate on a scale of 1-5 (where 5 is the best and 1 is the worst, also scores are only in integer values) and provide a brief explanation for each of the following criteria:} \\
\\
\texttt{1. \textbf{AUTHENTICITY}: How well does the recipe maintain the essential characteristics of the original dish?}
\texttt{Example: For a request to create a Korean-style spaghetti recipe, an extremely poor case would be generating a recipe for japchae.}\\
\texttt{2. \textbf{SENSITIVITY}: How well does the recipe understand and incorporate the target variation (Cuisine Transfer)?} \texttt{Example: For a request to create a halal version of spaghetti, an extremely incorrect case would be including pork as an ingredient.}\\
\texttt{3. \textbf{HARMONY}: How well does the generated recipe balance both AUTHENTICITY and SENSITIVITY? In other words, how well-crafted is the recipe overall?}
\\
\texttt{Format your response as follows:} \\
\texttt{AUTHENTICITY: <rating>} \\
\texttt{Reason: <brief explanation>} \\
\texttt{SENSITIVITY: <rating>} \\
\texttt{Reason: <brief explanation>} \\
\texttt{HARMONY: <rating>} \\
\texttt{Reason: <brief explanation>} \\
\end{minipage}}
\caption{Evaluation prompt given to models for each generated recipe. The prompt includes a scoring system (1-5 scale) and detailed explanations of the three evaluation criteria: \aut (Preservation of the original dish's characteristics), \sen (Accurate incorporation of the target cuisine), and \har (Overall balance between \aut and \sen. To ensure a structured evaluation result, a specific response format was also given. We obtained a total number of $129,600$ evaluation results.}
\end{figure*}

All LLM recipe evaluators were prompted with the same instruction format which is shown in Figure 5. Additionally, a group of five human annotators from diverse backgrounds (USA, South Korea, South Korea, Uganda, and Russia) were asked to evaluate a randomly selected set of $200$ recipes, stratified by cuisine. This process yielded an additional $1,000$ human-annotated ratings, bringing the total number of evaluation results to $130,600$.

\section{Evaluator Details}
\label{supple:evaluator_detail}
The models used for evaluating the recipes (recipe evaluators) are not only the open source LLMs previously used in the cuisine transfer task, but also large-scale proprietary LLMs (\texttt{gpt-4o}, \texttt{gemini-1.5-flash}, \texttt{gemini-1.5-pro}) as well.

Each recipe evaluator was instructed to rate the $4,800$ cuisine-transferred recipes each created by the six recipe generators, based on the above-mentioned criteria, along with a brief rationale that explains each of its decision. We repeated the evaluation process for each recipe generator three times.

\end{document}